\documentclass[acmsmall]{acmart}

\AtBeginDocument{%
  \providecommand\BibTeX{{%
    \normalfont B\kern-0.5em{\scshape i\kern-0.25em b}\kern-0.8em\TeX}}}

\setcopyright{acmcopyright}
\copyrightyear{2022}
\acmYear{2022}
\acmDOI{10.1145/3519296}

\acmJournal{JACM}
\acmVolume{37}
\acmNumber{4}
\acmArticle{111}
\acmMonth{2}

\usepackage{booktabs}
\usepackage{graphicx}
\usepackage{multirow}
\usepackage{algorithm}
\usepackage{algorithmic}
\usepackage{amsmath}
\begin{document}

\title{Deep Understanding based Multi-Document Machine Reading Comprehension}

\author{Feiliang Ren}
\email{renfeiliang@cse.neu.edu.cn}
\orcid{0000-0001-6824-1191}
\affiliation{%
	\institution{Northeastern University}
	\streetaddress{11 Wenhua Rd}
	\city{Heping Qu}
	\state{Shenyang City}
	\country{China}}

\author{Yongkang Liu}
\orcid{0000-0003-3098-0225}
\affiliation{%
	\institution{Northeastern University}
	\country{China}
}

\author{Bochao Li}
\orcid{0000-0003-2897-3886}
\affiliation{%
	\institution{Northeastern University}
\country{China}}
\author{Zhibo Wang}
\orcid{0000-0001-8170-2212}
\affiliation{%
	\institution{Northeastern University}
\country{China}}
\authornote{These authors contribute equally to this research and are listed randomly.}

\author{Yu Guo}
\orcid{0000-0003-4824-586X}
\affiliation{%
	\institution{Northeastern University}
\country{China}}
\authornotemark[1]
\author{Shilei Liu}
\orcid{0000-0003-2976-6256}
\affiliation{%
	\institution{Northeastern University}
\country{China}}
\authornotemark[1]
\author{Huimin Wu}
\orcid{0000-0001-9257-1342}
\affiliation{%
	\institution{Northeastern University}
\country{China}}
\authornotemark[1]
\author{Jiaqi Wang}
\orcid{0000-0001-9306-8757}
\affiliation{%
	\institution{Northeastern University}
\country{China}}
\authornotemark[1]
\author{Chunchao Liu}
\orcid{0000-0002-0028-8425}
\affiliation{%
	\institution{Northeastern University}
\country{China}}
\authornotemark[1]
\author{Bingchao Wang}
\orcid{0000-0002-3528-773X}
\affiliation{%
	\institution{Northeastern University}
\country{China}}
\authornotemark[1]

\renewcommand{\shortauthors}{Ren, et al.}

\begin{abstract}
Most existing multi-document machine reading comprehension models mainly focus on  understanding the interactions between the input question and  documents, but ignore following two kinds of understandings. First, to  understand the  {semantic meaning} of words in the input question and documents  from the perspective of each other. Second, to understand the supporting cues for a correct answer from the perspective of intra-document and inter-documents. Ignoring these  {two kinds}  of important understandings would make the models oversee some important information that may be helpful for finding correct answers. To {overcome this deficiency},  we propose a deep understanding based model for multi-document machine reading comprehension. It has three cascaded deep understanding modules which are designed to understand the accurate  {semantic meaning} of words, the interactions between the input  question and documents, and the supporting cues for  the correct answer. We evaluate our model on  two large scale benchmark datasets, namely TriviaQA Web and   DuReader. Extensive experiments show that our model  achieves state-of-the-art results on both  datasets.
\end{abstract}



\setcopyright{acmlicensed}
\acmJournal{TALLIP}
\acmYear{2022} \acmVolume{1} \acmNumber{1} \acmArticle{1} \acmMonth{1} \acmPrice{15.00}\acmDOI{10.1145/3519296}


\maketitle

\section{Introduction}
\label{section:intro}
Machine reading comprehension (MRC) aims to answer questions by reading given documents. %To this end, it demands the ability of well understanding natural language text and the ability of inference and reasoning. Thus, 
It is considered one of the core abilities of artificial intelligence (AI) and  the foundation of many AI-related applications like next-generation search engines and conversational agents. In real-world scenarios, MRC is often required to answer questions based on multiple documents. So multi-document MRC is receiving {growing} research {interests}~\cite{joshi2017triviaqa,clark2018simple,yan2019a,hu2019read,peng2020bi,zemlyanskiy2021readtwice}.
%Wang:18a,

{Generally}, there are {following} three main challenges in the multi-document MRC. (i) It requires  a model have the ability of processing very long text. For example, in {TriviaQA Web ~\cite{joshi2017triviaqa}, a benchmark multi-document MRC dataset,} 
there are averagely about 7 documents for each question {in its training set},  and each document contains averagely {about} 2,895 words. In DuReader \cite{he2018dureader}, another benchmark multi-document MRC dataset, there are about 5 documents for each question, and each document contains averagely {about}  1,793 Chinese characters. In contrast, in SQuAD \cite{rajpurkar2018know}, a benchmark single-document MRC dataset, there is only one document for each question, and each document contains averagely {about} 735 words.  (ii) In the multi-document MRC, there are many  {distractors} of an answer: some spans  have very high lexical matching results with the answer but  completely different semantic meaning with the answer. Thus it requires a model have the ability of accurately understanding the  {semantic meaning} of words in a document {and its  {corresponding}  question}.  (iii) The location of  an answer is very flexible in the multi-document MRC: it may appear once or multiple times in only a document, and it may also appear  multiple times in multiple documents. Obviously, this kind of information is useful for finding correct answers by {mutual authentication from following two aspects. (i) If a text span (not some meaningless function words) appears repeatedly in the input documents, it  would be highly possible to be related to the correct answers;  (ii)  If a text span only appears once in only a document, it would be less possible to be related to the correct answer.}. Thus   it requires a model have the ability of  mining such kind of information accurately. 

%T
Although these challenges are difficult to handle, researchers notice that human readers can well overcome them by using some reading patterns {like the patterns of ``\emph{read + verify}'' or \emph{multi-step reasoning}}. Inspired by this, {researchers}  begin to imitate human's reading patterns when they design  MRC models and lots of novel multi-documents MRC models are proposed~\cite{hu2019read,clark2018simple,yan2019a,wang2018multi,zhang2021retrospective,peng2020bi,chen2020multi,tian2020scene,malmaud2020bridging}.  Experiments show that these imitations  are very  effective and the  corresponding models achieve  state-of-the-art results on many benchmark datasets. 

%In fact, the underlying motivation of human readers using diverse reading patterns is to help them comprehensively \emph{understand} the semantic meanings of  the given documents and questions so as to find the correct answers accurately. However, most  existing methods pay more attention to these reading patterns' \emph{superficial frameworks}, but ignore some important \emph{understanding appeals}. 
{However, most of these  existing methods pay more attention to a reading pattern's \emph{superficial frameworks}, which means they are prone to design a model that have the same or similar processing  steps as a human's reading pattern. For example, if they imitate human's \emph{``read + verify"}   reading pattern, then they are prone to design a \emph{read} module and a \emph{verify} module in their MRC model. Similarly, if they simulate human's multi-step reasoning pattern, then they are prone to design an iterative-style MRC  model. The main deficiency of these  existing models is that they ignore  the underlying motivations of human readers using diverse reading patterns are  to  comprehensively \emph{understand} the  {semantic meaning} of  the given documents and questions.
	Some researchers \cite{zhang2020semantics,mihaylov2019discourse,guo2020a,gong2020hierarchical}  explore the semantic information understanding issue, but their methods either require some prerequisite resources like an extra knowledge base \cite{guo2020a} or the linguistic annotations \cite{mihaylov2019discourse}, or depend on some large scale pretrained language models~\cite{zhang2020semantics}.} %Besides, these methods pay more attention to a kind of coarse-grained understanding.}    

{We further notice that } there are {usually} three kinds of  { 
\emph{hierarchical understandings}} when human readers conduct {a} reading comprehension task, including the \emph{semantic meaning understanding of words}, the \emph{{interaction} understanding}  between the input question and documents, and the \emph{answer supporting {cue} understanding}   among different documents. 
Most existing models focus on designing attention based methods for  the \emph{{interaction} understanding} and designing a simple embedding layer for  the  \emph{semantic meaning understanding of words}, but paying less attention to the \emph{answer supporting {cue} understanding}.  We call these  existing methods as shallow  \emph{understanding} based models, and they usually suffer from  following two deficiencies. First, these models could not accurately \emph{understand} the   {semantic meaning}  of words. In the MRC task, the input question and documents are deeply correlated. Thus their words'  {semantic meaning}  should not be understood in isolation. Especially when the input question and documents contain  out-of-vocabulary (OOV) words, polysemy phenomenon,  and synonymy phenomenon.  
Second, these models do not make full use of the information provided by documents. Usually, a question's given documents have similar  {semantic meaning}, and the answer may occur in some of them or  appear many times in one of them. All  such information is helpful for finding the answer and should be fully used. 
%As  illustrated in Table \ref{tab:example}, although the word "\emph{Kilimanjaro}" is an OOV word, we can still accurately \emph{understand} its semantic when place it in the whole document. Furthermore,  many documents talk about something about "\emph{Mount Kilimanjaro}", which increases the probablity of it being the answer. In the multi-document MRC, each document describes something that is related to the question.
\begin{table*}
\centering
\caption{An example extracted from TriviaQA Web. The answer is  in \textbf{bold} and the key information is in \emph{italic}.}
\label{tab:example}

\begin{tabular}{lp{35em}}
	\hline
	Q     & {Which volcano in Tanzania s the highest mountain in Africa?} \\
	\hline
	A     & {\textbf{Mount Kilimanjaro} } \\
	\hline
	P1    & \textbf{Mount Kilimanjaro} , \emph{the Highest Volcano in Tanzania , Africa} | World Tourism Place \textbackslash{}n Stunning Views \textbackslash{}n \textbf{Mount Kilimanjaro} , \emph{the Highest Volcano in Tanzania , Africa}...is one of \emph{the highest volcanoes in the world and is the highest mountain in Africa}... \\
	\hline
	P2    & \textbf{Mount Kilimanjaro} - \emph{Tanzania Africa} - YouTube \textbackslash{}n ... Welcome to \textbf{Mount Kilimanjaro} a dormant volcano which is \emph{the highest mountain in Africa}...\emph{ in Tanzania , Africa}...\\
	\hline
	P3    & ... Sunrise on \textbf{Mount Kilimanjaro} . \textbackslash{}n © Anna Omelchenko/Fotolia \textbackslash{}n A caldera on Kibo , \textbf{Mount Kilimanjaro}...\\
	\hline
	P4    & ... Where is \textbf{Mount Kilimanjaro} \textbackslash{}n The cloud-swathed peaks of \emph{Africa ’ s highest mountain}... \\
	\hline
\end{tabular}

\end{table*}

To address these two deficiencies,  we propose a \emph{deep understanding} based multi-document MRC model. The core idea of our method can be briefly illustrated by the example demonstrated in  Table \ref{tab:example}. In this example, even if  ``\emph{Tanzania}" in the question is an OOV word, its semantic meaning can still be well understood when using the given documents as context since there is much key information available for understanding its accurate semantic meaning. For example, the context ``\emph{...the Highest Volcano in ...}'' and  ``\emph{...the highest mountain in ...}'' occur many times around ``\emph{Tanzania}", which indicates that ``\emph{Tanzania}" is highly possible to be  a location.   Besides, ``\emph{Mount Kilimanjaro}" occurs many times in a document and  many documents talk about  it, both  increase the probability of it being the answer. 

Specifically, the proposed  model contains three  {cascaded} deep understanding modules to imitate human's three kinds of {\emph{understandings}}. Besides the widely discussed \emph{{interaction} understanding}, our model {can also understand}: (i) the  {semantic meaning} of words by placing them into some specific contexts: taking documents as context when \emph{understanding} the semantic meaning of a word in the question, and taking the question as context when \emph{understanding} the semantic meaning of a word in documents, and (ii) the answer supporting cues by mining features from the aspects of intra-document and inter-document. 

We evaluate our model on two large-scale multi-document MRC benchmark datasets,  TriviaQA Web~\cite{joshi2017triviaqa} and  DuReader~\cite{he2018dureader}.  Extensive experiments  show that the proposed model is very effective and it achieves  competitive results on both of them.%\footnote{Due to the double-blind policy, here we do not provide the concrete ranks of our model on TriviaQA(Web) and DuReader test set leader-boards.}.  % On the deadline (June 1$^{st}$, 2020) of this submission\footnote{Our model achieved its best results on Decemer 2019, which are No.1 and No.3 on TriviaQA(Web) and DuReader test set leader-boards respectively.}, it ranks No.2 and No.6 on TriviaQA(Web) and DuReader test set leader-boards respectively. %It ranks No.1 on TriviaQA(Web) test data leaderboard and No.3 on DuReader test data leaderboard at the time of writing this paper(December 10$^{th}$, 2019).   

\section{Related Work}
According to the number of  documents given for a question, we categorize the MRC task into  single-document MRC and multi-document MRC. 

\subsection{Single-document MRC} 
\label{sec:single}
Based on the work of  ~\cite{seo2016bidirectional,yu2018qanet,nishida2019multi}, etc, we classify the main modules {in the models of this kind of MRC task } into following four layers. (i) \emph{Embedding layer} {that} aims to obtain an embedding representation for each word in the input question and documents. This layer can also be used to obtain the basic {semantic meaning} of a word, but it could not well address the common issues of OOV words, polysemy phenomenon, and synonymy phenomenon in natural language. Some researchers  integrate extra language models like  \emph{BERT}~\cite{devlin2018bert} or \emph{XLNet}~\cite{yang2019xlnet} into this layer, which can alleviate above issues but  {the cost is} introducing too many parameters. The models with large amount of parameters  require very large memory  {hardware}, which may be unaffordable to many users.  (ii)\emph{Matching layer} {that} is used for mining the interactions between the input question and documents.  It is often the core module in most existing MRC models and has been widely explored. Lots of attention based methods are proposed in this layer. For example, \emph{BiDAF}~\cite{seo2016bidirectional}  designs a context-to-query and query-to-context bi-directional attention method. Many other researchers, such as \cite{yu2018qanet,clark2018simple}, also use a \emph{BiDAF}-style attention method in this layer. Besides, \cite{cui2017attention} design an attention-over-attention method. \cite{wang2018multi} design a multi-granularity hierarchical attention method. \cite{yan2019a} use the self-attention method.  (iii) \emph{Model layer} {that} often uses \emph{LSTM} or \emph{CNN} based methods to capture the interactions among documents' words conditioned on the question features. 
(iv) \emph{Prediction layer} {that} often uses the \emph{pointer networks} to predict the probability of each position in the context being the start or end of an answer. 

It should be noted that the emergence of \emph{BERT}~\cite{devlin2018bert} and lots of its variants (like  \emph{XLNet}~\cite{yang2019xlnet}, \emph{RoBERTa}~\cite{liu2019roberta}, and \emph{ALBERT}~\cite{lan2020albert}, etc.)  greatly boost the benchmark performance of current MRC models due to their strong capacity for capturing the contextualized sentence-level language representations\footnote{ {In most pretrained language model based models, like the \emph{BERT}-based models, a separated token \emph{CLS} is often padded to the beginning of 	an input sentence, and its embedding representation is believed to contain the general information of the whole  input sentence,  and  is often used as a representation of this sentence.}}~\cite{zhang2021retrospective}. These language models  {simplify} the building of an MRC model and lots of most recent MRC models \cite{gong2020recurrent,zheng2020document,luo2020map,long2020synonym,banerjee2021self,li2020towards,zhang2020learn,guo2020incorporating,guo2020a,li2020mrc,huang2020nut,chen2020forcereader} only consist of   a language model based encoder module and an MRC task specific decoder module. However, there is a fatal deficiency for these language models. First, except \emph{XLNet},  \emph{BERT} and its other variants (\emph{ALBERT}, \emph{RoBERTa}, etc.) are auto-encoding based models, which  limits input size of  512 TOKENS~\cite{zemlyanskiy2021readtwice,gong2020recurrent}. This restriction has no effect on most {AI}-related   applications and most of single-document MRC tasks, but for {a} multi-document MRC dataset like DuReader or TriviaQA Web, this restriction will make most correct answers be excluded from the input documents even after a carefully designed data selection module. As for \emph{XLNet}, it is an auto-regressive based model, and can handle long text theoretically. However, it is an uni-directional model which can make predictions based on  forward information only, and can not use the backward information. %Thus, XLNet is more suitable for a generation task (like machine translation, dialogue generation, etc) other than a prediction task like MRC. 

%As ~\cite{zhang2021retrospective} point out that most MRC researchers keep the primary focus on the encoder side since they can simply benefit from a strong encoder. , and these language models . 

\subsection{Multi-document MRC} 
\label{sec:multi}
For this kind of MRC task, researchers  often design similar layers as in  the single-document MRC task but integrate new techniques to make full  use  of the multi-document information. Initially, researchers use simple reading strategies. For example, \cite{wu2018fast}  convert the multi-document data into the single-document format and then use single-document MRC models to find {answers. For example, } \cite{clark2018simple} first predict which paragraph to read and then apply models like \emph{BiDAF} to pinpoint the answer within that paragraph.  %For example, \cite{xu2019multi} truncate passage to contain at most 1000 tokens during training and eliminate those data with answers located after the $1000^{th}$ token.
Obviously, these simple methods {could not} make full use of information contained in the multi-documents, thus  researchers begin to design more sophisticated models to address the multi-document MRC task.  For example, \cite{wang2018multi} design three different modules in their  model, which can find the answer boundary, model the answer content, and perform cross-passage answer verification respectively. \cite{yan2019a} develop a novel deep cascade learning model that progressively evolves from the document-level and paragraph-level ranking of candidate texts to a more precise answer extraction.  \cite{xu2019multi} propose a multitask learning  model with a sample re-weighting scheme.% to assign sample-specific weights to the loss. 

Recently, the models of imitating reading patterns used by human are achieving more and more research attention due to their competitive results  on many benchmark MRC datasets.  For example,~\cite{sun2019improving} explicitly use three human’s reading strategies in their MRC model, including: (1) back and forth reading, (2) highlighting, and (3) self-assessment.
\cite{wang2018multipassage} imitate human's following reading pattern: first scans through the whole passage; then with the question in mind, detects a rough answer span; finally, come back to the question and select a best answer. \cite{liu2018stochastic} design their MRC model by simulating human's multi-step reasoning pattern: human  often re-read and re-digest given documents many times before a final answer is found. \cite{wang2018joint} use an extract-then-select reading strategy.  They further regard the candidate extraction as a latent variable and train the two-stage process jointly with reinforcement learning.  \cite{peng2020bi} design their MRC model by  simulating two ways of human thinking when answering questions, including reverse thinking and inertial thinking.    \cite{zhang2021retrospective} imitate human's \emph{``read + verify''}   reading pattern: first to read through the full passage along with the question and grasp the general idea, then re-read the full text and verify the answer. Some other researchers~\cite{hu2019read,clark2018simple,yan2019a,wang2018multi} also imitate human's \emph{``read + verify''}   reading pattern. Besides, there are other kinds of human reading patterns imitated, like the pattern of restoring a scene according to the text  {to} understand the passage comprehensively~\cite{tian2020scene}, the pattern of human gaze during reading comprehension~\cite{malmaud2020bridging}, the pattern of  tactical comparing and reasoning over candidates while choosing the best answer\cite{chen2020multi}, etc. %Some latest work also explores the 

Here we classify all these existing models  as a kind of {\em shallow understanding} based methods, since they pay more attention to these reading patterns' \emph{superficial frameworks}, but ignore some important {\emph{understandings}} hidden in these patterns.   
% For example,
\begin{figure}[tp]%%图
\centering  
\includegraphics[width=0.93\linewidth]{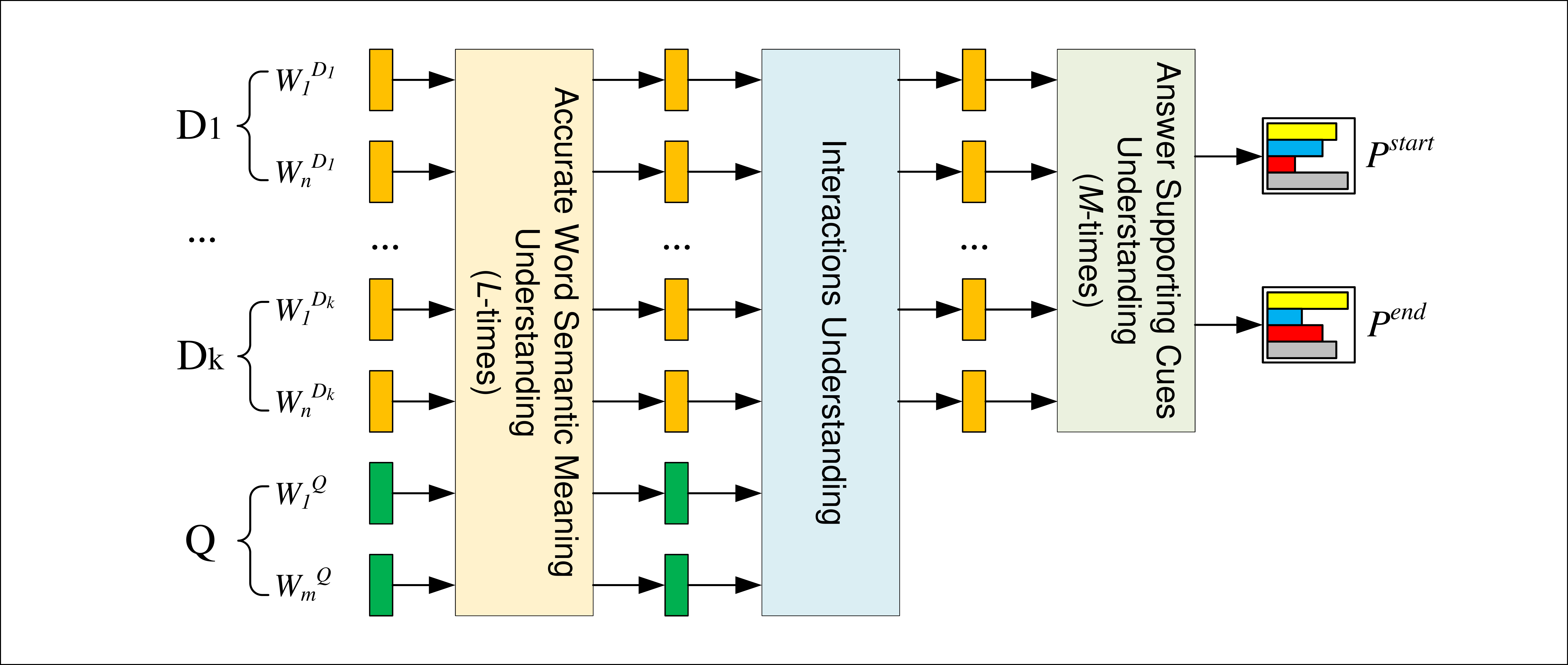}%{RimReader.png} 
\caption{The Framework of Our Model.}  
\label{fig:frame}%
\end{figure}
%Our method also imitates human's reading pattern, but it tries to imitate the \emph{deep understanding} process when human reading.  Accordingly, the characteristic of \emph{deep understanding}  distinguishes our model from other   existing  methods. %In a word, the characteristic of \emph{deep understanding}  distinguishes our
\section{Methodology}
\label{sec:method}
The framework of our model is shown in Figure \ref{fig:frame}. It mainly consists of  three \emph{understanding} modules that are designed to  imitate human's three kinds of  {\emph{understandings}} respectively.  

Given a question and some documents, the \emph{Accurate Word Semantic Meaning Understanding} module will generate a vector representation for each word in these input texts. These vector representations are expected to contain the accurate  {semantic meaning} of words when considering them in the overall context (including the question and the given documents). Then the \emph{{Interaction} Understanding} module further mines the interactions between the question and its documents, and outputs a new vector representation for each word in the documents. In each of these vector representations, the question-aware features are integrated. It should be noted that the input of this module includes all the vector representations that correspond to the {words} in both the question and its documents, but the output of this module only includes the vector representations that correspond to the words in the documents.  Next, taking these vector representations as input, the \emph{Answer Supporting {Cue} Understanding} module further mines the cue information which can indicate the possibility of a word in the documents being a context word in {an} answer. This module will output a refined vector representation for each word in the documents. Finally, based on the newest vector representations, the model computes the  probabilities of each word being the start and end {tokens of an answer}. Based on these probabilities, {an} answer can be deduced.

\subsection{Accurate Word Semantic Meaning Understanding}
\label{sec:accurate}
Accurately understanding the  {semantic meaning} of words in the input question and documents  is often the first and  most basic step when human readers solve {a} reading comprehension {task}. Accordingly, the aim of this module is to  {fulfill} this kind of understanding.  The embedding layer in traditional MRC models can achieve this aim to some extent. However, these  embedding based methods {could not} accurately understand a word's semantic meaning because they  generate a semantic representation for each word by only taking limited context (often the text where the word occurs) into consideration, which makes the generated representations are not  expressive enough especially when there are {phenomena} of OOV, polysemy and synonymy in the input text.
Some researchers \cite{yang2019enhancing,zhang2020sg,yang2019end} explore to integrate extra  {pretrained} language models in the embedding layer, which can alleviate above issues but at  
the cost of introducing more parameters.  \cite{dai2020multi} introduce a token-level dynamic reader to select important intermediate words according to boundary words, but they {do not} aim to  understand the  {semantic meaning} of words. %there is still much   room for improvement. 

We notice that  the input question and documents are often highly related to each other. Thus the input   question and documents can be viewed as the context of each other, which means the information from one part is useful to  understand the  {semantic meaning} of words in another part. For example,  in Table \ref{tab:example}, when  the word ``\emph{Tanzania}" in the question is an OOV word, it would be very difficult to \emph{understand} its semantic meaning when only considering the context of the question itself. However, it is still possible to obtain the expected semantic meaning when placing it in the given documents.  
Inspired by this observation, we design a \emph{coarse-to-accurate} method for \emph{word semantic meaning understanding}. Specifically, we first use a word embedding based method to obtain a word's shallow {semantic meaning}, then refine this meaning by integrating context information.  This is also in line with human  {readers'} reading pattern that they often grasp the literal meaning of a word firstly,  and then verify this  meaning  by placing it into different  contexts  {to} obtain the accurate semantic meaning of this  word in the given context.%The whole process is in line with human's reading pattern that  usually first obtain a word's shallow semantic from its basic meaning, then refine this semantic by placing it in a specific context. %The key of the \emph{fine} step is to find the accurate semantics possessed by words within a specific context. %In other word, the accurate  semantic of a word should be highlighted in the context.
%To this end, %it is natural to use an attention-based method. Inspired by this, 
%we design a cross-attention based method to obtain the accurate sematntics of words. 
%The whole process is also in line with the mentioned human's two-step understanding pattern. % reading pattern: they often first obtain the basic semantics of words, then refine the semantics with the help of context. 
%Besides, the refining step will be repeated \emph{L} times for this manner has been proven to be effective ~\cite{liu2018stochastic} for the MRC task. 

%Given a question and $k$ documents,  we adopt some widely used word embedding   techiuqes\footnote{Here we use word embedding, character embedding, and highway network. } to  generate $k+1$ word representation sequences for them. We use $h^Q=(h_1^Q,h_2^Q,...,h_m^Q)$ and $h^{D_t}=(h_1^{D_t},h_2^{D_t},...,h_n^{D_t})$ to denote the   sequences for the question $Q$ and its \emph{t-th}  document $D_t$ respectively. $m$ and $n$ are the numbers of words in the  question and the \emph{t-th} document.  We then refine the semantics by integrating context information to get the accurate word semantics.

\textbf{Coarse Word Semantic Meaning Understanding} Given a question and $k$ documents, we adopt some widely used word embedding  {techniques}\footnote{Here we use the GloVe word embeddings~\cite{pennington2014Glove}, the character embeddings that are generated by a common CNN model, and the highway network. All of them are widely used in existing MRC models.} to generate $k+1$ word representation sequences for them. We use $h^Q=(h_1^Q,h_2^Q,...,h_m^Q)$ and $h^{D_t}=(h_1^{D_t},h_2^{D_t},...,h_{n_t}^{D_t})$ to denote the sequences for the question $Q$ and its \emph{t-th}  document $D_t$ respectively, where $m$ and $n_t$ are the numbers of words in the  question and the \emph{t-th} document.  Each item in these sequences can be viewed as a \emph{coarse} semantic meaning for the corresponding word. Then these \emph{coarse}  {semantic meaning}  will be refined  by integrating context information to get the final \emph{accurate}  {semantic meaning}.

\textbf{Accurate Word Semantic Meaning Understanding} This step is expected to: (i) highlight the accurate  {semantic meaning} of words in documents from the perspective of the question, and (ii) highlight the accurate  {semantic meaning} of words in a question from the perspective of  documents. Obviously, this  {expectation} matches the principle of \emph{attention} mechanism well. Thus, here we design a \emph{cross attention} based method to obtain the accurate  {semantic meaning} of words in  the input question (or  documents) by taking  documents (or the  question) as context. Specifically, the designed methods has following four steps. 
%highlight the accurate semantics of words in documents from the perspective of a question and to highlight the accurate semantics of words in a question from the perspective of documents. To this end, it is natural to use a cross attention based method. The computation process of this cross attention  method is described as follows. 
%This module is designed in a  \emph{L}-hop manner to obtain the fine  semantics of words in a question within the context of documents and to obtain the fine  semantics of words in documents within the context of a question. 

\emph{Step 1:} we first compute a cross-attention matrix $A$, each of its element $A_{i,j} $ indicates the relevance between the \emph{i}-th word in $D_t$ and the \emph{j}-th word in \emph{Q}. $A$ is computed with Eq. \eqref{eq:co-matrix}. 
\begin{equation}
A_{i,j} = {(h_i^{D_t}})^T\mathbf{W}h_j^{Q}+\mathbf{U_l}\odot h_i^{D_t} +\mathbf{U_r}\odot h_j^{Q} 
\label{eq:co-matrix}
\end{equation}
where $\mathbf{W}$, $\mathbf{U_l}$ and $\mathbf{U_r}$ are trainable {matrices},  $\odot$ denotes the inter production operation, and in all of this paper, the superscript \emph{T} denotes a transpose operation.

%where the superscript \emph{T} means a transposition operation,  $W$ is a transformation matrix, % $W \in R^{d \times d}$, $U_l \in R^{d}$, $U_r \in R^{d}$, $d$ is the dimension of word representations in $D_t$ and $Q$, 
%and $\odot$ denotes the inter production operation. 

\emph{Step 2:} we assign an attention weight for each word in the input question and documents. And the attention weight  of a word is computed with Eq. \eqref{eq:att-weight}.
\begin{equation}
\begin{aligned}
	\alpha_i^{D_t}=softmax(A_{i:})\\ \alpha_j^{Q}=softmax(A_{:j}) 
	\label{eq:att-weight}
\end{aligned}
\end{equation}

\emph{Step 3:} based on the attention weights generated in previous step, we generate  $\tilde{h}_i^{D_t}$ and $\tilde{h}_j^{Q} $, which are new representations for a word in a document $D_t$  and {a word in} the question $Q$ respectively.  They are computed with Eq. \eqref{eq:att-rst}.
\begin{equation}
\tilde{h}_i^{D_t} =h^{Q}\alpha_i^{D_t}, \qquad \tilde{h}_j^{Q}=h^{D_t}\alpha_j^{Q} 
\label{eq:att-rst}
\end{equation}

\emph{Step 4:} we perform a bi-directional GRU based fusion operation to further refine the results generated in previous step, as shown in Eq. \eqref{eq:co-fusion1} and Eq. \eqref{eq:co-fusion2}, where $\mathbf{W_f^{(.)}}$ are trainable {matrices}. 
\begin{equation}
\begin{aligned}
	&f_i^{D_t}=[h_i^{D_t};h_i^{D_t}-\tilde{h}_i^{D_t};h_i^{D_t} \odot \tilde{h}_i^{D_t}]; \\ 
	&\tilde{f}_i^{D_t} = Relu(\mathbf{W_f^D}f_i^{D_t} + b_f); \\ 
	&\bar{f}_i^{D_t}=BiGRU(\tilde{f}_i^{D_t},\bar{f}_{i-1}) 
	\label{eq:co-fusion1}
\end{aligned}
\end{equation}
\begin{equation}
\begin{aligned}
	&f_j^{Q}=[h_j^{Q};h_j^{Q}-\tilde{h}_j^{Q};h_j^{Q} \odot \tilde{h}_j^{Q}]; \\
	&\tilde{f}_j^{Q} = Relu(\mathbf{W_f^Q}f_j^{Q} + b_f); \\
	&\bar{f}_j^{Q}=BiGRU(\tilde{f}_j^{Q},\bar{f}_{j-1}) 
	\label{eq:co-fusion2}
\end{aligned}
\end{equation}

The resulted $\bar{f}_i^{D_t}$ and $\bar{f}_j^{Q}$ denote the new  representations for the \emph{i-th} word in $D_t$ and the \emph{j-th} word in $Q$, {each of them} corresponds to a refined  semantic meaning of a word. %of which has a refined semantic.

As shown in Figure \ref{fig:frame}, the above four steps will be  iterated $L$ times  {to} obtain the final \emph{accurate} semantic meaning for each word. This repeated manner has been proven to be effective ~\cite{liu2018stochastic} for {an} MRC task. 
It should be noted that the matrices or vectors used in {above equations} are different for each iteration. For example, there will be $L$ different  $W$ in Eq. \eqref{eq:co-matrix}. Here for simplicity, we do not make distinction in the equation descriptions.  

%Besides, the refining step will be repeated \emph{L} times for this manner has been proven to be effective ~\cite{liu2018stochastic} for the MRC task. 
Finally, this  \emph{accurate word semantic meaning understanding} module outputs a new embedding representation for each word in the  question and {the} given documents. We denote the final word representation sequences for  $Q$ and $D_t$ as $\bar{f}^{Q}=(\bar{f}_1^{Q},\bar{f}_2^{Q},...,\bar{f}_m^{Q})$ and $\bar{f}^{D_t}=(\bar{f}_1^{D_t},\bar{f}_2^{D_t},...,\bar{f}_{n_t}^{D_t})$ respectively. And each item in these sequences can be viewed as the final \emph{accurate} semantic meaning for the corresponding word. 

%Finally, this  \emph{accurate word semantic understanding} module outputs a new embedding representation for each word in the  question and its documents. We denote the final word representation sequences for  $Q$ and $D_t$ as $\bar{f}^{Q}=(\bar{f}_1^{Q},\bar{f}_2^{Q},...,\bar{f}_m^{Q})$ and $\bar{f}^{D_t}=(\bar{f}_1^{D_t},\bar{f}_2^{D_t},...,\bar{f}_n^{D_t})$ respectively. 

\subsection{{Interaction} Understanding}

This module aims to find some important cues from the given documents that are helpful for locating an  answer. The difficulty for achieving this goal is how to accurately understand the interactions between the input question and documents. Different from the \emph{word semantic meaning understanding} that mainly focuses on the word-level  understanding, this module will focus on the document-level (or paragraph-level if we view each document and the question as a paragraph) {semantic meaning} understanding. 
We notice that when human readers solve this problem, they often first analyze the interactions between the input question and documents, then keep the question in mind and  re-read the documents to find the answer.  
%documents and questions, then keep questions in mind they often first obtain the rough interactions between a question and its  documents, then keep the question in mind and begin to re-read the documents to obtain the accurate interactions. 
Inspired by this, we design a \emph{two-step {interaction} understanding} method that is similar to human's above reading strategy. Specifically, it first analyzes the interactions between the input question and documents, then integrates the question features into the representations of words in documents  {to} form a question-aware representation for each  word in documents. 

\emph{Step 1:} in this step, it is a natural way to design a bi-directional attention based method due to the following two reasons. First, interactions are always bi-directional. Second, understanding interactions is to find which words are more helpful from the perspective of finding {an} answer, which is in line with the principle of attention mechanism. 

The attention method used in  \emph{BiDAF}~\citep{seo2016bidirectional} has been proven to be a very powerful method for understanding the interactions between the input question and documents and is widely used by lots of existing MRC models~\cite{yu2018qanet,clark2018simple}. Thus in this step, we  use the same attention method with \emph{BiDAF}. We omit the description of this attention computation  process  and readers can find the detail information in the original paper. Here we directly use $\{h^{D_t2Q}\} \in R^{n_t \times d}$ and $\{h^{Q2D_t}\} \in R^{m \times d}$ to denote the resulted document-to-question and question-to-document attended vectors, which are  {outputted} by the \emph{BiDAF} based method. %Here we directly use $\{h_k^{D2Q}\}_{k=1}^d$ and $\{h_k^{Q2D}\}_{k=1}^d$ to denote the resulted document-to-question and question-to-document attended vectors. 

\emph{Step 2:} in this step, we also use a BiDAF-alike fusion method to combine the  attention vectors and the embeddings obtained in previous \emph{word semantic meaning understanding} module together to yield a document representation sequence \emph{G}, each of its items $g^{D_t} \in R^{n_t * d}$ denotes  a new representation for a document $D_t$ where the question-aware information is integrated. But what's different with \emph{BiDAF} is that here we use a BiGRU based fusion function other than a simple concatenation operation. Specifically, $g^{D_t}$ is computed with Eq. \eqref{eq:bi-rst}.
%\begin{small}
\begin{equation}
\begin{aligned}
	g_i^{D_t}=BiGRU(g_{i-1}^{D_t},[\bar{f}_i^{D_t};h_i^{D2Q};\bar{f}_i^{D_t} \odot h_i^{Q2D}; \bar{f}_i^{D_t} \odot h_i^{D2Q}]) 
	\label{eq:bi-rst}
\end{aligned}
\end{equation}
%In the second step, the above attention vectors and the embeddings obtained in previous \emph{word semantic understanding} module are combined together to yield $\{g_k^{D_t}\}_{k=1}^m$, a new representation for a document $D_t$ that the question-aware information is integrated. $g_k^{D_t}$ is computed with Eq. \eqref{eq:bi-rst}.

%\begin{small}
%	\begin{equation}
%		\begin{aligned}
	%			g_k^{D_t}=BiGRU(g_{k-1}^{D_t},[\bar{f}^{D_t};h_k^{D2Q};\bar{f}^{D_t} \odot h_k^{Q2D}; \bar{f}^{D_t} \odot h_k^{D2Q}]) 
	%			\label{eq:bi-rst}
	%		\end{aligned}
%	\end{equation}
%\end{small}

It should be noted that as shown in Figure \ref{fig:frame}, we do not perform a repeated operation in this module. {This is because that the {input} of this module {contains} word representations of both  the input question and documents, but the {output} only {contains} the word representations of documents. Thus the number of input tokens is different from the number of output tokens. 
Of course, a linear transformation can be used to map the {output} of this module to the same size as the {input}. But this would be lack of a reasonable explanation: in the original {input}, each representation correlates with a real word either in the question or in its documents, { but it is very difficult to ask  the transformed results still can be semantically correlated  with these input words. In fact, our subsequent experiments show that repeating this module with a linear transformation operation  is  much harmful to the performance of our model.}}

% Thus there is a risk that after several repeated operations, the  {semantic meaning} of the transformed results are far and far away from those of the input words, which would be 
%{This is because that the  embeddings of the question tokens will have no valuable information added: since the purpose of this intersection module is to compare each word from each document to question, which allows the model to select the best document.}

% due to the following reason. The {input} of this module {contains} word representations of both  the input question and documents, but the {output} only {contains} the word representations of documents. Thus the number of input tokens is different from the number of output tokens. 
%Of course, a linear transformation can be used to map the {output} of this module to the same size as the {input}. But this would be lack of a reasonable explanation: in the original {input}, each representation corresponds to a true word either in the question or in its documents, { but it is very difficult to ask  the transformed results maintain the same or similar  {semantic meaning} with these input words. Thus there is a risk that after several repeated operations, the  {semantic meaning} of the transformed results are far and far away from those of the input words, which would be  much harmful to the performance of our model.}

\subsection{Answer Supporting {Cue} Understanding}
In the multi-document MRC, every document is expected to contain the answer or some information that is highly related to the answer, thus the  {semantic meaning} of different documents would be highly related to each other.  Accordingly, if an answer candidate in a document is the correct answer, it would be highly possible to achieve extra supporting cues\footnote{Here we define the answer supporting cues as a kind of  information that is very helpful for locating an answer.} from other documents. Besides, the word representations {generated} by previous \emph{{interaction} understanding}  module are question-aware, so a correct answer would also be highly possible to achieve extra supporting cues from words in the same document. 
%As  illustrated in Table \ref{tab:example},  many documents talk about "\emph{Mount Kilimanjaro}" and  "\emph{Mount Kilimanjaro}" also frequently occurs in a document, which increases the probablity of it being the answer. 
%In other word, the correct answer would be  possible to achieve supporting cues from both intra-document and inter-document. Obviously, both  kinds of supporting cues are suitable for being understood by an attention based method. % for they match the principle of the \emph{attention} mechanism.
Based on these analyses, we design an \emph{intra-document} and \emph{inter-document} self-attention based method to collect these supporting cues. 

\textbf{Intra-document Answer Supporting {Cue} Understanding} is a self-attention based method that is designed to highlight  the answer's content words  from the perspective of other words in the same document where these content words appear. In other words, this module is expected to highlight some words that are regarded as answer words by  {most words}.  
Specifically, it generates $f^{D_t} \in R^{n_t * d}$, a new word representation sequence for each  document $D_t$, as shown in Eq. \eqref{eq:intra-att}.
\begin{equation}
f_i^{D_t}=BiGRU(f_{i-1}^{D_t},[g_i^{D_t},w_i])
\label{eq:intra-att}
\end{equation}
where $g_i^{D_t}$ is the representation of the \emph{i}-th token in the document $D_t$ and is {generated} by previous \emph{{interaction} understanding} module,  $w_i$ is the attention value between $g_i^{D_t}$ and $g^{D_t}$, and is computed with a widely used attention computation method \cite{wang2017gated,bahdanau2015neural} as shown in Eq. \eqref{eq:intra-attV}, where $\upsilon$ is a trainable vector, $\mathbf{W^{D_t}}$ and $\mathbf{V^{D_t}}$ are trainable {matrices}.
\begin{equation}
\begin{aligned}
	&s_i^j=\upsilon^Ttanh(\mathbf{W^{D_t}}g_i^{D_t}+\mathbf{V^{D_t}}g_j^{D_t}), \\
	&\alpha_i^j=exp(s_i^j)/\sum\limits_{j=1}^{n_t}exp(s_i^j),   \\ &w_i=\sum\limits_{j=1}^{n_t}\alpha_i^jg_j^{D_t}  
	\label{eq:intra-attV}
\end{aligned}
\end{equation}

\textbf{Inter-document Answer Supporting {Cue} Understanding} is design to highlight  the answer's content words  from the perspective of other documents. In other words, this module is expected to highlight some words that are regarded as answer words by  {most  documents}.  Specifically, for each document, we first concatenate all its words' representations obtained in previous \emph{intra-document supporting {cue} understanding} step together to form a new representation for this document. Accordingly, we will obtain a new document representation sequence $P=\{f_1^{D_1},f_2^{D_1},...,f_{1}^{D_k},...,f_{n_t}^{D_k}\}$, and each item in this sequence corresponds to the representation of a document. 
Then the inter-document self-attention is performed on $P$ to generate $F_p=\{f_p^1,f_p^2,...,f_p^L\}$, where $L=\sum_{i=1}^{k}n_i$. Each of its item  $f_p^i$ corresponds to the representation of a word in the concatenated document, and is computed with the method show in Eq. \eqref{eq:inter-att} and  \eqref{eq:inter-attV}.  
\begin{equation}
f_p^i=BiGRU(f_p^{i-1},[P_i;\beta_i])
\label{eq:inter-att}
\end{equation}
\begin{equation}
\begin{aligned}
	&s_i^j=\gamma^Ttanh(\mathbf{W_f}P_i+\mathbf{V_f}P_j),   \\  &\mu_i^j=exp(s_i^j)/\sum_{j=1}^Lexp(s_i^j) , \\ 
	&\beta_i=\sum_{j=1}^L\mu_i^jP_j 
	\label{eq:inter-attV}
\end{aligned}
\end{equation}
where $\mathbf{W_f}$ and $\mathbf{V_f}$ are trainable {matrices}, and $\gamma$ is a trainable vector.

As shown in Figure \ref{fig:frame}, the \emph{answer supporting {cue} understanding}  {module} will be  repeated $M$ times so that more accurate supporting cues are highlighted. 
Finally, we still denote the output of this module as   $F_p$, each item of which corresponds to a word representation where different {kinds of} \emph{understanding} information is integrated.

\subsection{Answer Prediction}
\label{sec:ans}
We  use a pointer networks based method that is similar to the ones   in BiDAF~\cite{seo2016bidirectional} and Match-{LSTM}~\cite{wang2017machine} to predict the probability of each word in $F_p$ being the start  or the end  of an answer span. The pointer networks \cite{vinyals2015pointer}  produce only the start token and the end token of {an} answer, and then all the tokens between these two {tokens} in the original passage are considered to be the {correct} answer. 
Specifically, the probability distributions of the start and end indexes over {tokens of  all  documents} are computed with Eq.\eqref{eq:prob}. % The pointer networks produces only the start token and the end token of the answer,  which can be formulated as follows:
\begin{equation}
\begin{aligned}
	&\mathbf{P}^s = \beta_s^Ttanh(\mathbf{W_p^s}F_p + \mathbf{W_g^s}G) \\
	&\mathbf{P}^e = \beta_t^Ttanh(\mathbf{W_p^e}F_p + \mathbf{W_g^e}G) 
	\label{eq:prob}
\end{aligned}
\end{equation}
where $\mathbf{W_p^{(.)}}$ and $\mathbf{W_g^{(.)}}$ are trainable {matrices}, $\beta_{(.)}$ are trainable vectors, and  \emph{G} is the output of  {the} previous \emph{ {interaction} understanding} module. {Note that \emph{G} has the same number of tokens as  $F_p$}.

Finally, we define the loss function as the negative sum of the log probabilities of the predicted distributions indexed by the true start and end indices over all samples, as shown in Eq. \eqref{eq:loss}. 
\begin{equation}
Loss=-\sum\limits_{i}^N[log(p_{y_i^s}^b)+log(p_{y_i^e}^e)]
%Loss=-\frac{1}{N}\sum\limits_{i}^N[log(p_{y_i^b}^b)+log(p_{y_i^e}^e)]
\label{eq:loss}
\end{equation}
where $y_i^b$ and $y_i^e$ are the true start and end index of the \emph{i-th} sample respectively.

At the inference time, an answer candidate $A_i^{'}$ (we denote its start and end indices as \emph{x} and \emph{y} respectively) is chosen with the maximum value of $a_x^ba_y^e$ under a constraint that $x\leq y$. 

%\subsection{Data Preprocessing}
%\label{section:data}
%In our model, we also design a simple string matching based data pre-processing module to filter out some irrelevant sentences from each question's given documents. Specifically, we compute a \emph{cosine} similarity between each sentence of input documents and the ANSWER\footnote{In our in-house experiments, we find that using \emph{answer}  achieves better experimental results than using “question”.}. Only the sentences whose similarities are higher than a predefined threshold would be left for model training. %During testing, for each question, we take a simple  truncating strategy that only leaves 500 words\footnote{The number “500” is setted based on the experinental results on validation set.} from its  documents.

\section{Experiments}

\subsection{Datasets and Experimental Settings}

We evaluate our model on TriviaQA Web ~\cite{joshi2017triviaqa} and DuReader~\cite{he2018dureader}, two large-scale multi-document MRC benchmark datasets. %in real-world settings.

\noindent\textbf{TriviaQA} is an English MRC dataset containing over 650K question-answer-evidence triples. It includes 95K question-answer pairs authored by trivia enthusiasts and independently gathered evidence documents, six per question on average, which are generated from either  Wikipedia or Web search. \textbf{Note} there are  two separated datasets in TriviaQA: one is TriviaQA Wiki which is for {the} single-document MRC, and the other is TriviaQA Web which is for {the} multi-document MRC. In TriviaQA Web, besides the full development and test set, a verified subset for each is also provided.

\noindent\textbf{DuReader} is a Chinese multi-document MRC dataset, which is designed to address real-world MRC. It 
has three advantages over previous MRC datasets. First, all of its questions and documents are based on Baidu Search and Baidu Zhidao, and the answers are manually generated. Second, it provides rich annotations for more question types. Third, it is a large MRC dataset that contains 200K questions, 420K answers and 1M documents. 

\noindent\textbf{Implementation Details} \label{part:detail} In experiments, the dimension of word embeddings and the hidden layer in the \emph{BiGRU} unit are set to 300 and 150 respectively. During training, word embeddings are not updated and the batch size is set to 16. Adam optimizer is used and the learning rate is set to 0.001. Training epoch is set to 2. On DuReader test set leader-board, ROUGH-L and BLEU4 are used as evaluation metrics. On TriviaQA Web test set  leader-board, EM and F1 are used as evaluation metrics. In experiments, the ensemble model is obtained by averaging 4 single models’ prediction probabilities. 
DuReader provides   \emph{free-form} reference answers that  not all   can be found in the input documents. So for each question, as the method used in ~\cite{wang2018multipassage}, we choose the span that achieves the highest ROUGE-L score with its reference answers as the {golden} span for training. % and

{During training, we also design a simple string matching based data  {preprocessing} module to filter out some irrelevant sentences from each question's given documents. Specifically, we compute a \emph{cosine} similarity between each sentence of input documents and the ANSWER\footnote{In our in-house experiments, we find that using \emph{answer}  achieves better experimental results than using “question”.}. Only the sentences whose similarities are higher than a predefined threshold would be left for model training. During  {testing}, answers are not available,  we select the sentences by computing the \emph{cosine} similarity between sentences in the input documents and the questions.}

\noindent\textbf{Baselines} Following  strong state-of-the-art  models are taken as baselines: BiDAF~\cite{seo2016bidirectional}, Smartnet~\cite{chen2017smarnet}, Fast\cite{wu2018fast}, Simple~\cite{clark2018simple}, QANet~\cite{yu2018qanet}, Cascade~\cite{yan2019a}, Match-LSTM~\cite{wang2017machine}, R-Net~\cite{wang2017gated}, 	PR+BiDAF{~\cite{wang2018multi}},
CrossPassage{~\cite{wang2018multipassage}, and BCTN{~\cite{peng2020bi}}. All of them are the best  multi-document MRC models that can be found so far\footnote{It should be noted that there are some models that appear on the test set leader-boards of some MRC  datasets, but we {could not} find their corresponding papers either in  conferences, journals, or on arXiv.}.  Here except the results of BCTN, all the results of other  baselines are directly copied from \cite{yan2019a}. Besides, we also report the results of two popular {pretrained} language models: one is \emph{RoBERTa}~\cite{liu2019roberta}, and the other is \emph{XLNet}~\cite{yang2019xlnet}. Both of them are {recent variants}  of \emph{BERT} and are reported to be superior to \emph{BERT} or other kinds of language models like \emph{Elmo}~\cite{peters2018deep} {on a lot of AI-related tasks}. As a  strong variant of \emph{BERT},  \emph{RoBERTa} can use the \emph{sliding window} based method to handle the text that is longer than 512 tokens\footnote{The used \emph{RoBERTa} model is implemented by a transformer code base, which can be found at following website: https://github.com/huggingface/transformers/.}. Besides. it should be noted that both  language models have two versions: base and large. {Here} we report their results of both versions. 
%DuReader ,  we choose the span that achieves the highest ROUGE-L score with the reference answers as the gold span for training. This method is also used in ~\cite{wang2018multi}.  
\begin{table}[t]
	\centering
	\caption{Effect of repeated numbers on DuReader.}
	\label{tab:exp1}
	
	%	\makebox[\linewidth]
	{\begin{tabular}{lllll}
			\hline
			\multicolumn{1}{c}{\multirow{2}[2]{*}{}} & \multicolumn{4}{c}{ROUGE-L/BLEU4} \\
			\multicolumn{1}{c}{} & \textit{\ \ \ M=1} & \textit{\ \ \ M=2} & \textit{\ \ \ M=3} & \textit{\ \ \ M=4} \\
			\hline
			\textit{L=1} & 53.12/50.74 & \textbf{54.23}/50.52 & 52.41/50.39 & 50.03/47.87 \\
			\textit{L=2} & 52.32/48.29 & 52.83/49.03 & 51.44/45.27 & 47.58/48.69 \\
			\textit{L=3} & 50.26/49.39 & 51.02/\textbf{53.28} & 46.17/48.24 & 45.66/47.19 \\
			\textit{L=4} & 48.13/48.28 & 48.69/50.81 & 47.32/49.29 & 44.33/44.75 \\
			\hline
	\end{tabular}}
\end{table}%
\begin{table}[t]
	\centering
	\caption{Main results on DuReader.  $^*$ indicates the results are generated by us. $^\dagger$ indicates that the results  are copied from \cite{peng2020bi} directly.}%
	\label{tab:main1}
	{\begin{tabular}{lll}
			\hline
			\multicolumn{1}{r}{} & \multicolumn{1}{p{4.5em}}{ROUGE-L} & \multicolumn{1}{p{3em}}{BLEU4} \\
			\hline
			MatchLSTM{~\cite{wang2017machine}} & 39    & 31.8 \\
			BiDAF{~\cite{seo2016bidirectional}} & 39.2  & 31.9 \\
			R-Net{~\cite{wang2017gated}} & 47.71 & 44.88 \\
			PR+BiDAF{~\cite{wang2018multi}} & 41.81 & 37.55 \\
			CrossPassage{~\cite{wang2018multipassage}} & 44.18 & 40.97 \\
			CascadeModel{~\cite{yan2019a}} & 50.71 & 49.39 \\
			\textit{XLNet-Base{{~\cite{yang2019xlnet}}}} &57.36$^*$ &49.21$^*$  \\
			\textit{XLNet-Large{{~\cite{yang2019xlnet}}}} &61.05$^*$ &54.38$^*$  \\
			RoBERTa-Base{~\cite{liu2019roberta}} & 54.18$^\dagger$ & 38.85$^\dagger$ \\
			RoBERTa-Large{~\cite{liu2019roberta}} & 59.12$^\dagger$ & 44.53$^\dagger$ \\
			BCTN-Base{~\cite{peng2020bi}} & 58.04 & 43.19 \\
			BCTN-Large{~\cite{peng2020bi}} & 59.12 & 44.53 \\
			
			\hline
			\textit{OurModel (Single)} & \textbf{62.19} & \textbf{56.34} \\
			\textit{OurModel (Ensemble)} & \textbf{63.36} & \textbf{57.91} \\
			\hline
	\end{tabular}}%
\end{table}%
\begin{table}
	\centering
	\caption{Main results on TriviaQA Web. $^*$ indicates the results are generated by us. }%
	\label{tab:main2}
	
	{\begin{tabular}{lll}
			\hline
			\multirow{2}[2]{*}{Model} & \ \ \ \ Full  & Verified \\
			\multicolumn{1}{c}{} & \ \ \ EM/F1 & \ \ EM/F1 \\
			\hline
			BiDAF{~\cite{seo2016bidirectional}} & 40.74/47.05 & 49.54/55.80 \\
			Smarnet{~\cite{chen2017smarnet}} & 40.87/47.09 & 51.11/55.98 \\
			Fast{~\cite{wu2018fast}} & 47.77/54.33 & 57.35/62.23 \\
			Simple{~\cite{clark2018simple}} & 66.37/71.32 & 79.97/83.70 \\
			QANet{~\cite{yu2018qanet}} & 51.1/56.6 & 53.3/59.2 \\
			Cascade{~\cite{yan2019a}} & 68.65/73.07 & 82.44/85.35 \\
			
			RoBERTa-Base{~\cite{liu2019roberta}} & 64.97$^*$/70.89$^*$ & 78.41$^*$/83.03$^*$ \\
			RoBERTa-Large{~\cite{liu2019roberta}} & 66.65$^*$/72.39$^*$ & 79.84$^*$/84.49$^*$ \\
			\textit{XLNet-Base{{~\cite{yang2019xlnet}}}} & 63.92$^*$/67.42$^*$ & 77.39$^*$/79.57$^*$  \\
			\textit{XLNet-Large{{~\cite{yang2019xlnet}}}} & 65.64$^*$/69.40$^*$ & 79.58$^*$/82.08$^*$  \\
			\hline
			\textit{OurModel (Single)} & \textbf{68.72/73.13} & \textbf{82.70/85.35} \\
			\textit{OurModel (Ensemble)} & \textbf{69.64/73.80} & \textbf{83.36/85.66} \\

			\hline
	\end{tabular}}%
\end{table}%
\subsection{Main Experimental Results}
\label{sec:exp}
{Because we do not have high performance (like large-memory or fast speed) GPU servers, } we first use 30,000 training samples and 6,000 testing samples of DuReader to quickly find the proper settings of \emph{L} and \emph{M}. Then we report all the other experiments based on these two fixed parameters. 
%We don't think this will change our conclusions because there is usually a similar performance trend between the models trained with smaller and larger samples. 
The results are shown in Table \ref{tab:exp1}, from which we can see that %\emph{L} and \emph{M} have different effects on ROUGE-L and BLEU4. We also notice that  
ROUGE-L and BLEU4 do not increase synchronously. %: sometimes ROUGE-L increases but BLEU4 decreases, or vice versa. 
This is mainly because  the provided answers in DuReader is \emph{free-form}, and we need to convert these reference answers into  spans to {fit in with} the  span extraction based methods. During this process, ROUGE-L is used as the guide metric. %Thus the extracted spans are inherent bias toward  ROUGE-L. 
So it  is  possible to make  ROUGE-L and BLEU4 reach their peaks under different conditions, which will then cause the mentioned phenomenon. In fact, this phenomenon is also common on other \emph{free-form} MRC datasets like MS MARCO~\cite{nguyen2017ms}. %Of course, if the provided answers are spans, both these metrices will reach the peaks synchronously.  %The best BLEU4 is obtained when \emph{L} is set to 3 and \emph{M} is set to 2. But the best ROUGE-L is obtained when smaller \emph{L} is set. 
Since DuReader test set leader-board takes ROUGE-L as the main evaluation metric, all the experiments are reported under the settings of $L=1$ and $M=2$ {where the model achieves the highest ROUGE-L score}.

%The main experimental results are summarized in Table \ref{tab:main1} and \ref{tab:main2}. Here the results of the baselines are directly taken from \cite{yan2019a}. From these results we can draw the conclusion that the proposed model is a very effective multi-document MRC model. On the TriviaQA(Web) test set leader-board,  our model achieves the \textbf{BEST} results\footnote{\url{https://competitions.codalab.org/competitions/17208\#results}. Our model’s ID is “\emph{NEUKG}”, we submitted our results on Aug 12, 2019.} up to now. On the DuReader test set leader-board, our model ranks No.3\footnote{\url{https://ai.baidu.com/broad/leaderboard?dataset=dureader\&task=Main} Our model’s ID is “\emph{RIMReader}” , we submitted our results on Dec 10, 2019} at the time of writing the paper(Dec. 10th 2019).

The main experimental results are summarized in Table \ref{tab:main1} and \ref{tab:main2}. From these results we can see that our model is  very effective: on both datasets and under all evaluation metrics,  it consistently outperforms all the compared state-of-the-art baselines.   %and it performs well on both  datasets. %Before this submission, the best results achieved by our method are No.1 on TriviaQA(Web) test set leader-board and No.3 on DuReader test set leader-board. %that At the time of writing this paper(December $10^{th}$, 2019), it ranked No.1 %\footnote{\url{https://competitions.codalab.org/competitions/17208\#results}, our model’s ID is “\emph{NEUKG}”.}  
%on TriviaQA(Web) test set leader-board and ranked No.3 %\footnote{\url{https://ai.baidu.com/broad/leaderboard?dataset=dureader\&task=Main}, our model’s ID is “\emph{RIMReader}”.} 
%on DuReader test set leader-board .

Furthermore, we can see that on both datasets, our model achieves much better results than both \emph{RoBERTa} and \emph{XLNet}.  We argue this is mainly due to following reasons. For \emph{RoBERTa}, its \emph{sliding window} mechanism can alleviate the issue of handling long documents.  This mechanism slices a document into multiple segments, and each segment will be individually encoded by {the  encoder}, finally all the encoded results of these  segments are merged. This will lead to following fatal deficiency. In this mechanism, each slice is encoded separately, which will lose much of important correlation information among documents.  {Especially} when the answer length exceeds 512 tokens, this mechanism will make the  {semantic meaning} of different {slices} incomplete, which is very harmful for finding some important cues  from {either} the intra-document level {or} the inter-document level. This deficiency will harm the performance greatly. 
As for  \emph{XLNet}, although  it can handle long text due to its  auto-regressive mechanism, its uni-directional processing property still makes it can not make full use of the given documents due to the {lose} of backward information. %from different perspectives.    

% Table generated by Excel2LaTeX from sheet 'Sheet3'

\subsection{Ablation Experiments}
To demonstrate the contributions of different components in our model, we conduct ablation experiments and the results are shown in Table \ref{tab:abl1}.
%The  ablation experiments of our metod are shown in Table \ref{tab:abl1} and \ref{tab:abl2}.

\noindent\textbf{Effectiveness of Accurate Word Semantic Meaning Understanding} From Table \ref{tab:abl1} we can see that when the \emph{accurate word semantic meaning understanding} module {is} removed, both  ROUGE-L and BLEU4 drop sharply on DuReader. %by a large margin  
Similar results can be seen on TriviaQA Web. These results show that it is helpful for accurately understanding the   {semantic meaning} of words in the input question and documents. 

To further evaluate the effectiveness of the \emph{accurate word semantic meaning understanding} module, we replace it with \emph{XLNet}. In other words, we use \emph{XLNet} on top of the common word embedding layer {since this practice is taken}  by lots of existing models.  New experiments are shown in Table \ref{tab:xlnet1} (Here the large version of \emph{XLNet} is used.). %From the results 
We can see that our designed \emph{word semantic meaning understanding module} works better. We think this is mainly due to following two reasons. First, as analyzed above, \emph{XLNet} can not make full use of the given documents due to the {lose} of backward information. Second,  \emph{XLNet} is a  {pretrained} model, some words'  {semantic meaning} generated by it may not well match the true scenario in an MRC dataset. One may argue that the parameters in \emph{XLNet} can be re-trained on a specific application scenario. But training large-scale language models is often very time-consuming and the required hardwares (such as GPU servers) are far beyond what we can afford. 

Besides, we can see that when \emph{XLNet} is integrated into our model, it  achieves much better results on both datasets than its original version.  These results indicate that {our} proposed model has a general framework that can be used to further boost the performance of existing MRC models.
%Besides, we can see that compared with the original version, 
%In \emph{Appendix}, we compare the numbers of paramenters between with and without \emph{XLNet} in our model. 

%We do not compare the performance of using other language models like \emph{BERT} mainly for the following reasons. For all the public available \emph{BERT} we can find, the input text length is restricted to 512 TOKENS~\cite{zemlyanskiy2021readtwice,gong2020recurrent}. This restriction has no effect on most \emph{AI}-related   applications and some simple MRC tasks. But for either DuReader or TriviaQA Web, this restriction will make most correct answers be excluded from the input documents even after a carefully designed data selection module. %Also, retraining a new \emph{BERT} model is far beyond what we can afford.

% Table generated by Excel2LaTeX from sheet 'Sheet4'
\begin{table}[t]
	\centering
	\caption{Ablation experiments on TriviaQA Web (\emph{upper part}) and DuReader (\emph{bottom part}).}
	\label{tab:abl1}%
	{\begin{tabular}{lll}
			\hline
			\multirow{2}[2]{*}{\textit{Model}} & \ \ \ \ Full  & Verified \\
			\multicolumn{1}{r}{} & \ \ \ EM/F1 & \ EM/F1 \\
			\hline
			\textit{OurModel(Single) } & 68.72/73.13 & 82.70/85.35 \\
			\textit{-CrossAttU} & 67.19/71.01 & 77.86/82.45\\
			\textit{-InteractionU} & 64.21/68.68 & 75.47/78.17  \\
			\textit{-SupportingCueU} & 60.12/62.97& 71.32/72.45 \\
			\textit{-Intra-DocSelfAtt} & 68.38/72.62 & 81.92/84.69 \\
			\textit{-Inter-DocSelfAtt} & 66.87/71.22 & 79.68/83.87 \\
			\textit{-DataPreprocessing} & 66.43/71.35 & 79.97/83.70  \\
			\hline
			Model & \multicolumn{1}{l}{ROUGE-L} & \multicolumn{1}{l}{BLEU4} \\
			\hline
			\textit{OurModel(Single) } & 62.19 & 56.34 \\
			\textit{-CrossAttU} & 61.37 & 55.04 \\
			\textit{-InteractionU} & 58.78 & 54.17 \\
			\textit{-SupportingCueU} & 54.30 & 49.21 \\
			\textit{-Intra-DocSelfAttn}  & 60.25 & 55.43 \\
			\textit{-Inter-DocSelfAtt} & 58.78 & 54.39 \\
			\textit{-DataPreprocessing} & 60.04 & 54.46 \\
			\hline
	\end{tabular}}%
	
\end{table}%

\begin{table}[t]
	\centering
	\caption{Comparisons of with/without \emph{XLNet} on TriviaQA Web (\emph{upper part}) and DuReader (\emph{bottom part}).} %\emph{XLNet} has two versions: base and large. In all our experiments, we use the large version.}
\label{tab:xlnet1}%

{\begin{tabular}{lll}
		\hline
		\multirow{2}[2]{*}{\textit{Model}} & \ \ \ \ Full  & Verified \\
		\multicolumn{1}{r}{} & \ \ \ EM/F1 & \ EM/F1 \\
		\hline
		\textit{OurModel(Single) } & 68.72/73.13 & 82.70/85.35 \\
		%			\textit{XLNet{{~\cite{yang2019xlnet}}}} & 65.64/69.40 & 79.58/82.08  \\
		\textit{OurModel+XLNet} & 66.09/69.78 & 79.71/81.88 \\
		%			\hline
		\midrule
		\multicolumn{1}{r}{} & \multicolumn{1}{l}{ROUGE-L} & \multicolumn{1}{l}{BLEU4} \\
		\hline
		\textit{OurModel(Single)} & {62.19} & {56.34} \\
		%			\textit{XLNet{{~\cite{yang2019xlnet}}}} &61.05 &54.38  \\
		\textit{OurModel+XLNet} &61.50  &54.79  \\
		\hline
\end{tabular}}%
\end{table}%


%Intra-document Answer Supporting Cues Understanding
\noindent\textbf{Effectiveness of Answer Supporting {Cue} Understanding} We can see that when the whole \emph{answer supporting {cue} understanding} module {is} removed, the performance drops sharply. But when either the \emph{intra-document} or {the} \emph{inter-document answer supporting {cue} understanding} module used, the model achieves competitive results. Besides, the performance drops more when the \emph{inter-document answer supporting {cue} understanding} module {is} removed, which shows the supporting cues from other documents play more roles than that of from a document itself. This is just like a voting process: the more documents provide supporting cues, the more likely an answer candidate  be the correct answer.  

\noindent\textbf{Effectiveness of  {{Interaction} Understanding}} From Table \ref{tab:abl1} we can see that the  \emph{{interaction} understanding}  modules are important. 

 {In fact, }  our model   is adaptable to different choices other than \emph{BiDAF} in  the \emph{{Interaction} Understanding} module. To evaluate this adaptability, we  use the interaction methods in  several other MRC models to replace \emph{BiDAF}, and the results are shown in Table \ref{tab:inter}. % other similar models like \emph{BiDAF} can also be used in the   \emph{interactions understanding} module. To evaluate the adaptation of this framework, we 
We can see that all these models have  similar  {contributions} as  \emph{BiDAF}. Furthermore,  from  Table \ref{tab:inter} we can see that when a model is integrated into the framework of our model, it always achieves much better results than its original version. Taking Match-LSTM as example, when it is used in our model, its results on both datasets are far higher than those of its original version. These results confirm again that the proposed model has a general framework and can be used to further boost the performance of existing MRC models. % draw the conclusion that the proposed framework is very effective and it  can  be used to improve the performance of other models.   % other similar models like \emph{BiDAF} can also be used in the   \emph{interactions understanding} module. To evaluate the adaptation of this framework, we 

 {Besides,  we also conduct experiments  that perform  a repeated operation in this \emph{interaction understanding} module by a simple linear transformation operation. The results are shown in Table ~\ref{tab:rep} (the results are obtained under our \emph{single} version model). We can see that there is a significant performance drop when this module begins to repeat (\emph{N} = 1). Then, as the the repeated number increases, the performance of our model drops accordingly. Especially, when \emph{N} = 3, the performance of our model is even worse than that of removing the whole \emph{interaction understanding} module. We take the sample shown in Table~\ref{tab:example} as a specific case, and use a simple \emph{inner-product} based method to compute the similarity between the original representation of the word ``\emph{Tanzania}" and its transformed representation. The  results show that the similarities between these two representation become lower and lower as the repeated number increases. Such results confirms our previous analyses that: in the original {input}, each representation correlates with a real word either in the question or in its documents,  but it is very difficult to ask  the transformed results still can be semantically correlated  with these input words. Thus there is a risk that after several repeated operations, the {semantic meaning} of the transformed results are far and far away from those of the input words, which would be  much harmful to the performance of our model. }  %Of course, a linear transformation can be used to map the {output} of this module to the same size as the {input}. But this would be lack of a reasonable explanation: in the original {input}, each representation corresponds to a true word either in the question or in its documents, { but it is very difficult to ask  the transformed results maintain the same or similar 
%each representation correlates with a real word either in the question or in its documents, { but it is very difficult to ask  the transformed results still can be semantically correlated  with these input words.
	%  {semantic meaning} with these input words. Thus there is a risk that after several repeated operations, the  {semantic meaning} of the transformed results are far and far away from those of the input words, which would be  much harmful to the performance of our model.} }
\begin{table}[t]
\centering
\caption{ {Comparisons} of using different \emph{{interaction} understanding} methods on TriviaQA Web (\emph{upper part}) and DuReader (\emph{bottom part}).} %\emph{XLNet} has two versions: base and large. In all our experiments, we use the large version.}
\label{tab:inter}%
{\begin{tabular}{lll}
	\hline
	\multirow{2}[2]{*}{\textit{Model}} & \ \ \ \ Full  & Verified \\
	\multicolumn{1}{r}{} & \ \ \ EM/F1 & \ EM/F1 \\
	\hline
	\textit{+BiDAF(OurModel)} & 68.72/73.13 & 82.70/85.35 \\
	\textit{+MatchLSTM{\cite{wang2017machine}}} & 66.54/72.28 & 80.33/84.12  \\
	\textit{+AOA{\cite{cui2017attention}}} & 66.77/72.45 & 81.31/84.89  \\
	\textit{+RNet{\cite{wang2017gated}}} & 67.94/72.42 & 81.22/84.97 \\
	
	\midrule
	\multicolumn{1}{r}{} & \multicolumn{1}{l}{ROUGE-L} & \multicolumn{1}{l}{BLEU4} \\
	\hline
	\textit{+BiDAF(OurModel)} & 62.19 & 56.34 \\
	\textit{+MatchLstm{\cite{wang2017machine}}} & {61.42} & {54.73} \\
	\textit{+AOA{{~\cite{cui2017attention}}}} &61.05 &54.38  \\
	\textit{+RNet{\cite{wang2017gated}}} &61.31  &55.74  \\
	\hline
\end{tabular}}%
\end{table}%
\begin{table}[t]
\centering
\caption{ {Effect of repeated numbers (\emph{N}) for the \emph{interaction understanding} module on DuReader and TriviaQA.} }
\label{tab:rep}

{\begin{tabular}{lllll}
	\hline
	\multicolumn{1}{c}{\multirow{2}[2]{*}{}} & \multicolumn{4}{c}{DuReader (ROUGE-L/BLEU4)} \\
	\multicolumn{1}{c}{} & \textit{\ \ \ N=0} & \textit{\ \ \ N=1} & \textit{\ \ \ N=2} & \textit{\ \ \ N=3} \\
	\hline
	&{62.19}/{56.34}  & {60.37}/{55.92} & {58.26}/{54.12} & {53.87}/{48.53} \\
	
	\hline
	
	\multicolumn{1}{c}{\multirow{2}[2]{*}{}} & \multicolumn{4}{c}{TriviaQA Full (EM/F1)} \\
	\multicolumn{1}{c}{} & \textit{\ \ \ N=0} & \textit{\ \ \ N=1} & \textit{\ \ \ N=2} & \textit{\ \ \ N=3} \\
	\hline
	&68.72/73.13  &  66.51/70.34 & 63.14/66.25 & 59.67/64.28 \\
	\hline	
	
	\multicolumn{1}{c}{\multirow{2}[2]{*}{}} & \multicolumn{4}{c}{TriviaQA Verified (EM/F1)} \\
	\multicolumn{1}{c}{} & \textit{\ \ \ N=0} & \textit{\ \ \ N=1} & \textit{\ \ \ N=2} & \textit{\ \ \ N=3} \\
	\hline
	& 82.70/85.35 & 80.04/82.14 & 77.57/72.69 & 73.09/68.24 \\
	\hline	
	
\end{tabular}}
\end{table}%
\begin{table}[t]
\centering
\caption{ {Comparisons} of parameter number (\emph{millions})  on TriviaQA Web and DuReader.}% for training, \emph{hours} for testing). }%
\label{tab:paras}

{\begin{tabular}{lll}
\hline
\multicolumn{1}{r}{} & \multicolumn{1}{l}{TriviaQA} & \multicolumn{1}{l}{DuReader} \\
\hline
%			\multicolumn{1}{r}{} & \multicolumn{2}{c}{ParameterNumber} \\
%			\hline
MatchLSTM{\cite{wang2017machine}} & {$\approx$128}  &{$\approx$93} \\% & 93,889,705    & 115,935,978\\
BiDAF{\cite{seo2016bidirectional}} & {$\approx$113}  &{$\approx$84} \\
\textit{XLNet{{~\cite{yang2019xlnet}}}}& {$\approx$146}  &{$\approx$123} \\% &123,066,627 &146,188,746  \\
\textit{OurModel+XLNet} & {$\approx$243}  &{$\approx$212}\\% &212,251,839  &243,977,502  \\
\textit{OurModel(Single)} & {$\approx$126} & {$\approx$92}   \\%& {$\approx$92,852,603} & {126,939,876} \\
\hline
%			\multicolumn{1}{r}{} & \multicolumn{2}{c}{TrainingTime} \\
%			\hline
%			MatchLSTM{\cite{wang2017machine}} & {$\approx$13h}  &{$\approx$9h} \\% & 93,889,705    & 115,935,978\\
%			BiDAF{\cite{seo2016bidirectional}} & {$\approx$12h}  &{$\approx$8h} \\
%			\textit{XLNet{{~\cite{yang2019xlnet}}}}& {$\approx$15h}  &{$\approx$10h} \\% &123,066,627 &146,188,746  \\
%			\textit{OurModel+XLNet} & {$\approx$26h}  &{$\approx$18h}\\% &212,251,839  &243,977,502  \\
%			\textit{OurModel(Single)} & {$\approx$19h} & {$\approx$11h}   \\%& {$\approx$92,852,603} & {126,939,876} \\
%			\hline
\end{tabular}}%
\end{table}%

\subsection {Parameter Efficiency} 
We  quantitatively compare the parameter numbers  of several models whose source codes are available. All the models are trained on a TitanRTX 8000  GPU server (\emph{XLNet} requires so large memory that it  couldn't be trained on a server like TitanXP) with the configurations that lead to the best performance we achieved. The comparison results are shown in Table \ref{tab:paras}. % less parameters, which means it can be trained on  a smaller memory server. In fact, our model can be trained on a TitanXP GPU server but \emph{XLNet} cannot. 
We can see that our model has less parameters than most of the compared models.   When taking the performance into consideration, we can conclude that  our model is more parameter efficient: it achieves  better results with fewer parameters.

Here we do not compare the  {run time} of different models {because} it is  difficult to provide a fair evaluation environment: coding tricks, hyper-parameter settings (like  \emph{batch-size}, \emph{learning rate}, etc), parallelization, lot of non-model factors  affect the  {run time}.

\subsection {Error Analyses} 
\label{section:err}
Here we make some error analyses. Specifically, on DuReader,  we randomly select 2,000 poorest ROUGE-L results {generated} by our model as error samples. And on  TriviaQA Web, we take all the results whose  EM values are wrong  on the development set as error samples. Then we try to classify these error samples into different groups according to their error types, and the results are shown in Table \ref{tab:errors}, in which all the listed examples are taken from TriviaQA Web. For clarity, we omit the given documents of each example since these documents on either dataset are very long.  % from two datasets. 

Generally,  there are following three main kinds of errors  on both datasets. (i) \emph{incomplete}, which means that only partial of a predicted answer matches the corresponding golden answer. (ii) \emph{redundant}, which means that a golden answer is a word subset of the  predicted answer. (iii) \emph{unanswerable}, which means that the question is unanswerable (an ideal model should identify  these  unanswerable questions  and refuse to give an answer for it), but the model outputs an answer. %And the first two kinds of errors can be unified as a kind of \emph{boundary recognition} errors, which means  the predicted answers have wrong boundaries compared with the golden answers. 

\begin{table}[t]
\centering
\caption{Error Analyses. \emph{Q} refers to \emph{question}.}% for training, \emph{hours} for testing). }%
\label{tab:errors}
\begin{tabular}{lllp{20em}}
\hline
\multicolumn{1}{l}{\multirow{2}[4]{*}{Error Types}} & \multicolumn{2}{c}{Proportion(\%)} & \multicolumn{1}{c}{\multirow{2}[4]{*}{Examples}} \\
\cmidrule{2-3}          & \multicolumn{1}{l}{DuReader} & \multicolumn{1}{l}{TriviaQA Web} & \multicolumn{1}{c}{} \\
\hline
incomplete  & 21.7  & 6.14  & \textbf{Q}: A Long Island Iced Tea is a cocktail based on vodka, gin, tequila, and which other spirit?\newline{}\textbf{Golden answer}: light rum\newline{}\textbf{Predicted answer}: rum \\
\midrule
redundant  & 35.5  & 23.68  & \textbf{Q}: What does a costermonger sell?\newline{}\textbf{Golden answer}: fruit\newline{}\textbf{Predicted answer}: fruit and vegetables \\
\hline
unanswerable  & 0.3   & 10.53  & \textbf{Q}: "Which US president was behind ""The Indian Removal Act"" of 1830, which paved the way for the reluctant and often forcible emigration of tens of thousands of American Indians to the West?"\newline{}\textbf{Golden answer}: null\newline{}\textbf{Predicted answer}: President Monroe \\
\hline
others & 42.5  & 59.65  & \textbf{Q}: Romaine \& Butterhead are types of what?\newline{}\textbf{Golden answer}: iceberg lettuce\newline{}\textbf{Predicted answer}: lettuces \\
\hline
\end{tabular}%

\end{table}%

On DuReader, both the \emph{incomplete} and the \emph{redundant}  {kind} of errors account for a large proportion of {all the errors}. And the \emph{other}  {kinds} of errors includes  \emph{partial matching} errors, \emph{yes/no} errors, etc. 
On TriviaQA Web,  the \emph{redundant} {kind} of errors account for a large proportion of {all the errors}, followed by the \emph{unanswerable} kind of errors. The \emph{unanswerable} kind of errors  account for significantly larger proportion on TriviaQA Web than on DuReader because there are far less \emph{unanswerable} kind of question on DuReader than that on TriviaQA Web.  {The \emph{other} kind of errors on TriviaQA Web include  errors }  like the \emph{named entity recognition} errors, \emph{singular and plural} errors, \emph{partial matching errors}, etc.
After detailed analyses of these errors we find that in most cases, the locations of the predicted answers are very close to the golden answers. In fact, these errors could be corrected only when {a} model do  \emph{understand} the main {semantic meaning} of the input text, which further indicates the reasonability of our research line. %during the MRC task.

\begin{figure}[t]%%图
\centering  
\includegraphics[width=\linewidth]{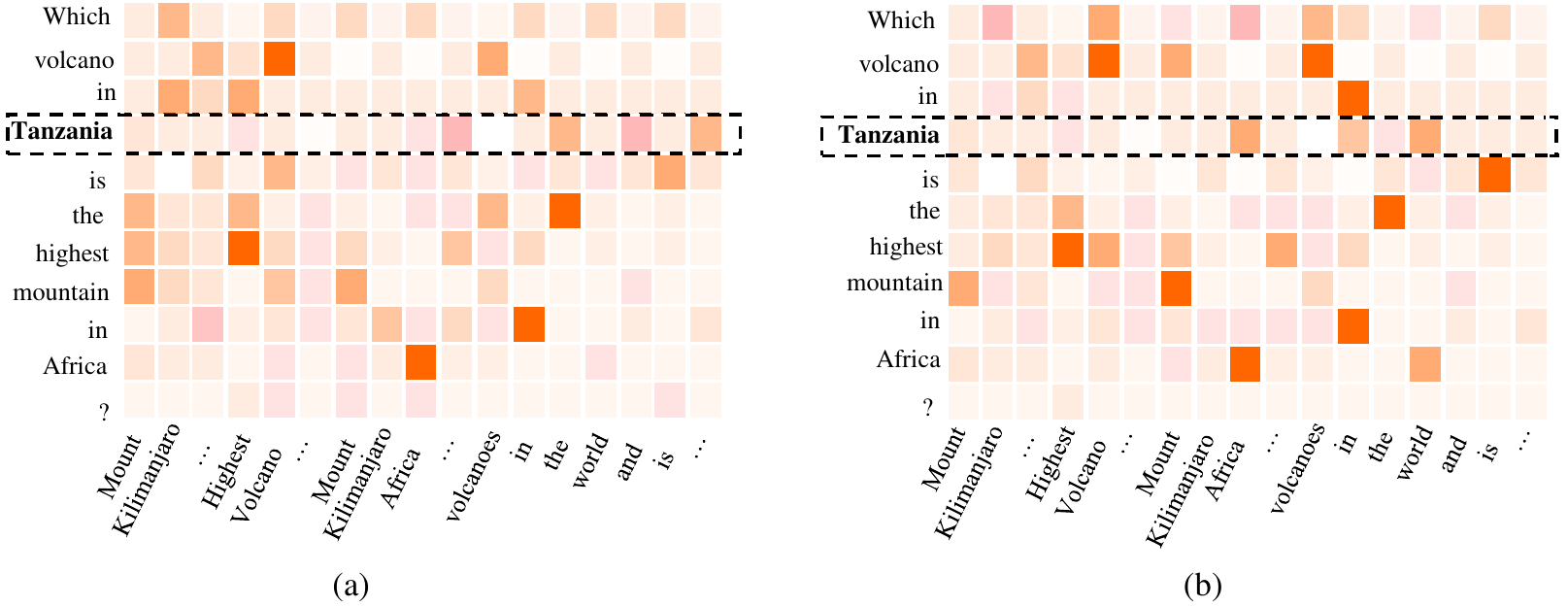}%{RimReader.png} 

\includegraphics[width=\linewidth]{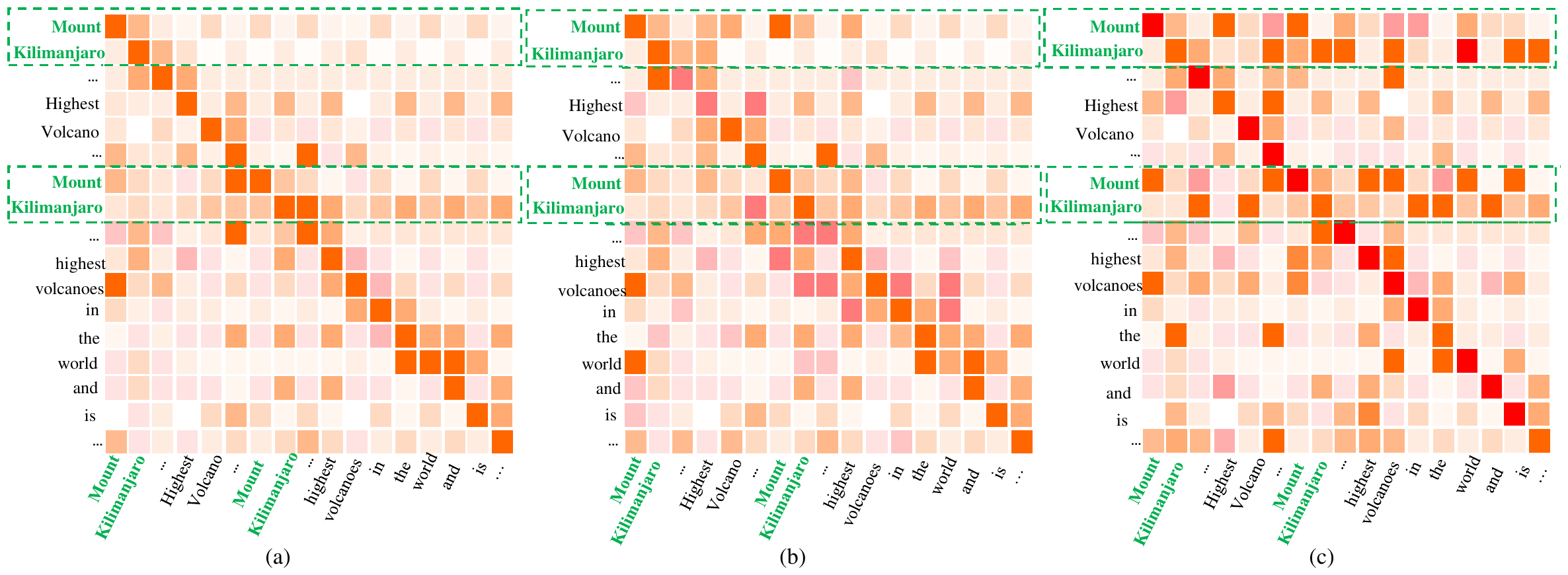}
\caption{\textbf{Upper}:  {comparisons} of the \emph{cosine} similarities between words' embeddings of before (\emph{subfigure a}) and after (\emph{subfigure b}) using the proposed ``\emph{Accurate Word Semantic Meaning Understanding}" module. \textbf{Bottom}:  {comparisons} of the attention weights when  using ``\emph{Intra-document Answer Supporting {Cue} Understanding}" (\emph{subfigure a}), ``\emph{Inter-document Answer Supporting {Cue} Understanding}" (\emph{subfigure b}), and both (\emph{subfigure c}). 
}  
\label{fig:frame1}%
\end{figure}

%\subsection{A Case Study}
\subsection {Case Study} 
Taking the  question and documents illustrated in Table \ref{tab:example} as an example, here we use Figure \ref{fig:frame1} to further demonstrate the effectiveness  of the proposed two \emph{understanding} modules. From the upper part of Figure \ref{fig:frame1} we can see that when  the proposed ``\emph{Accurate Word Semantic {Meaning} Understanding}" module used, ``\emph{Tanzania}" (in the question) obtains higher similarities with  words like ``\emph{Africa}" and ``\emph{world}", which highlights its accurate semantic {meaning} greatly. From the bottom part of Figure \ref{fig:frame1} we can see that the answer ``\emph{Mount Kilimanjaro}" achieves different attention weights when using different  \emph{answer supporting {cue} understanding}  {sub-module}. When {all the modules} used,  it achieves the highest  attention weight, which increases the probability of it being the answer.

\subsection{Discussions}

Before this submission, the best results achieved by our model  on TriviaQA Web and DuReader test set leader-boards were No.1\footnote{{https://competitions.codalab.org/competitions/17208\#results}, tab the \emph{Web} button.} and No.3\footnote{{https://ai.baidu.com/broad/leaderboard?dataset=dureader\&task=Main}}  respectively\footnote{Now it ranks No.2 and No.7 on these two test set leader-boards respectively.}. One can  notice that currently, most top models on different MRC leaderbords (like SQuAD\footnote{https://rajpurkar.github.io/SQuAD-explorer/}, HotPotQA\footnote{https://hotpotqa.github.io/}, CoQA\footnote{https://stanfordnlp.github.io/coqa/}, MS MARCO\footnote{https://microsoft.github.io/msmarco/}, etc.) depend heavily on  large scale  {pretrained} language models like \emph{BERT} (or its variants). However, these language models based MRC models have two fatal deficiencies.

First, they can only be run on high-cost hardware environments since  the language models  have so large amount of parameters that much large GPU memories are often required  to load these parameters. This will bring heavy {burdens} on most researchers since the cost of building such environments are very high. Accordingly, this will prohibit these models to be used on some real-time or online scenarios.  

Second, they can only be used on the scenarios where the maximum length of the input text is within a specific threshold since most existing language models like \emph{BERT} (including most of its variants) have a length restriction on the input text. This condition is not always met, especially for some languages like Chinese where the corresponding MRC task usually involves very long text. One may argue that this deficiency can be addressed by re-training a new language model. However, re-training such a new large scale language model without length restriction is far beyond the affordability of most researchers due to the high hardware requirement and the high time cost.   

Both deficiencies prohibit the adaptability of the language model based MRC models. On the contrary, our model is a simple and effective MRC model, and it has following  two overwhelming advantages compared with the language model based MRC models.  

First, our model uses simple technologies like GRU but achieve very competitive results, which means it can be very easily reproduced by other researchers. 

Second, {\cite{wang-etal-2020-tplinker} have pointed out in their work that for all  systems that use some  {pretrained} language models like \emph{BERT}, the language model  is usually the most time-consuming part and takes up the most of model parameters. In contrast,} our model does not use any  {pretrained} language models, thus {compared with the models that use per-trained language models, our model usually has a smaller parameter size and faster inference speed,}  which means it can well fit in with some online or real-time applications without {the requirements of} high-performance  {hardware}.

In a word, our model shows that by well \emph{understanding} the  { semantic meaning} of the input text,  the state-of-the-art performance still can be achieved  even without using  sophisticated technologies, high-cost  {hardware}, and large scale language models. %can be achieved  even with a simpler model framework.

%to improve the performance of  MRC, the keys are not only using sophisticated technologies,
%if the input text was well understood,  the state-of-the-art performance can be achieved  even with a simpler model framework. 

\section{Conclusions}
In this paper, we propose a simple but effective \emph{deep understanding} based  multi-document MRC model. It  uses neither any sophisticated technologies nor any  {pretrained} language models. We evaluate our model on DuReader and TriviaQA Web, two widely used benchmark multi-document MRC datasets.  Experiments show that our model achieves very competitive results on both datasets. %Besides, our model is very efficient since . 

The main novelties of our work are as follows. 
First,  our model   has a general  {framework} that consists of three understanding modules that imitate  human's three kinds of {understandings} during reading comprehension. Second, the designed \emph{accurate word semantic meaning understanding} module can well  \emph{understand} a word's  semantic meaning. It even  plays a better role than an extra language model like \emph{XLNet} but with far less parameters.  This  {is} very important for MRC's application to the online or real-time environments. 
Third, the designed  \emph{answer supporting {cue} understanding} module is effective, and it can   increase the probability of finding answers. %Besides, our model provides a general and adaptive framework, and either of its three key understanding modules can be updated by better  methods. %boost the reliabilities  of located answers. % being located by understanding \emph{intra-document} and \emph{inter-document} answer supporting cues.   

\begin{acks}
{This work is supported by the National Natural Science Foundation of China (No.61572120) and the Fundamental Research Funds for the Central Universities (No.N181602013).}
\end{acks}

\bibliographystyle{ACM-Reference-Format}
\bibliography{sample-base}


\end{document}